\documentclass[conference]{IEEEtran}
\IEEEoverridecommandlockouts
\usepackage{amsmath,amssymb,amsfonts}
\usepackage{algorithmic}
\usepackage{graphicx}
\usepackage{textcomp}
\usepackage{xcolor}
\usepackage[backend=bibtex, maxbibnames=99, firstinits]{biblatex}
\usepackage{pgfplots}
\usepackage{mathtools}
\usepackage{multirow}
\usepackage{adjustbox}
\def\BibTeX{{\rm B\kern-.05em{\sc i\kern-.025em b}\kern-.08em
    T\kern-.1667em\lower.7ex\hbox{E}\kern-.125emX}}
\begin{document}

\title{What augmentations are sensitive to hyper-parameters and why?}

\author{\IEEEauthorblockN{ Ch Muhammad Awais}
\IEEEauthorblockA{\textit{Machine Learning} and \\ \textit{Knowledge Representation} Lab \\
\textit{Innopolis University}\\
Innopolis, Russia \\
c.awais@innopolis.university}
\and
\IEEEauthorblockN{BEKKOUCH Imad Eddine Ibrahim}
\IEEEauthorblockA{\textit{Sorbonne Center for Artificial} \\ \textit{Intelligence - SCAI} \\
\textit{Sorbonne University}\\
Paris, France \\
imad.bekkouch@etu.sorbonne-universite.fr}
}

\maketitle

\begin{abstract} We apply augmentations to our dataset to enhance the quality of our predictions and make our final models more resilient to noisy data and domain drifts. Yet the question remains, how are these augmentations going to perform with different hyper-parameters? In this study we evaluate the sensitivity of augmentations with regards to the model's hyper parameters along with their consistency and influence by performing a Local Surrogate (LIME) interpretation on the impact of hyper-parameters when  different augmentations are applied to a machine learning model. We have utilized Linear regression coefficients for weighing each augmentation. Our research has proved that there are some augmentations which are highly sensitive to hyper-parameters and others which are more resilient and reliable.
\end{abstract}


\section{Introduction}
Machine Learning (ML) is the process of solving a particular problem using a computer based on the input data without being explicitly programmed. An ML model can be thought of as a function, mostly parametric, that maps the input (predictors) to output (class labels) and which is learned from the data using an optimization procedure \cite{issam2015ml}. ML falls into different categories such as, supervised, unsupervised, semi-supervised, reinforcement learning etc. Some applications of ML are, super market analysis \cite{4195245}, accurately counting footsteps by a step counting machine \cite{regression}, human activity recognition \cite{ANNad}, \cite{ANN2}, \cite{ANN3}, \cite{ANN4}, \cite{ANN5}, \cite{SVM1}, \cite{cHA}, image processing \cite{HRCF1}, \cite{HRCF2}, \cite{HRCF3}, \cite{HMM1}, \cite{HMM2}, \cite{HMM3}, \cite{MEMM}, and \cite{SVM2}, text analysis \cite{text} etc.
\newline\newline Deep Learning(DL)\cite{9228891} is a subcategory of ML, which consists of algorithms containing multiple processing layers that are self-capable of handling unstructured or unlabeled data \cite{dl2015yann}, based on hierarchy of data \cite{onExpert}. DL models learn the features of the data at every layer, and those layers are trained iteratively. Some of the DL applications include feature extraction and car accident detection using 3D Convolutional Neural Networks \cite{CNN3D1, 8936124, 10.1007/978-3-030-82196-8_50}, \cite{CNN3D2}, Classification of XRay, and other medical images \cite{deepLearningXRay, lens.org/035-746-864-097-526}, Hyperspectral Image classification using Auto-encoders \cite{AE1,10.1145/3440084.3441180} \cite{DL3}, and Hockey game monitoring \cite{skmo}, etc.
\newline\newline Unlike model parameters, which are learned from the data, ML also involves parameters that are external to the model and are used to control the learning process. Such parameters are called hyper-parameters. Some examples of hyperparameters are as follows:
\begin{itemize}
    \item Number of epochs: It defines the number of times the model should repeat the training for learning.
    \item Optimizer: It is the component that determines how the parameters of an ML model should change or update.
    \item Learning Rate: It determines how much the parameters should change while minimizing the loss function. We can use a large learning rate or a small learning rate. A large learning rate will speed up the process, but it might miss the local minimum, whereas a small learning rate can slow down the process. The optimum learning rate helps in reaching an optimum solution in less time. \cite{DBLP:journals/corr/abs-2007-15745}
    \item Batch size: It is the number of samples used in one iteration of the learning process.
    \item Test-Train-Valid Ratio: It determines the percentage of data we are going to use for testing, training, and validation of an ML model, respectively.
\end{itemize}
These hyper-parameters are not learnt from the data, and the values are estimated on the bases of different experimentation. 
\newline\newline After hyper-parameters, the next essential part of training an ML model is the data itself. If we have less data the model will not give us good results. Here comes the role of data augmentation, which deals with the automatic generation and expansion of data, which can lead to better learning of machines, resulting in models that can generalize better on unseen data \cite{CNN3D2, deepLearningXRay, segmentation, VAE1}. We will discuss data augmentations in more detail in section \ref{sec-LR}.

It is important to note that the performance of an ML model changes, if we change the hyper-parameters \cite{probst2018tunability} or if we change the type or combination of applied augmentations. Several studies have explored these aspects, but it will be interesting and important to study how these two (augmentations and hyper-parameters) work with each other. Accordingly, the aim of this work is to study the relationship between the augmentations and hyper-parameters but on a limited scope. Our hypothesis is that there are some augmentations which are hyper-parameter sensitive, i.e., they change their behavior when we apply different hyper-parameters on the same dataset and the same model. 

To test our hypothesis we used lime analysis, and linear-regression coefficients. We have studied the effect of nine different augmentations and two machine learning models with four different hyperparameters. We will elaborate it further in section \ref{sec:methods}. 

\section{Literature Review}\label{sec-LR}
Data Augmentations can be divided into two main categories, pre-processing augmentations and in-processing augmentations. The pre-processing augmentations part is done before feeding data to the ML model and processing it to increase the size of dataset. In this paper our focus is mainly on image augmentations, some of the pre-processing techniques for image augmentation are as follows:
\subsection{Data Augmentation Techniques}
\subsubsection{Geometric Augmentations}
\begin{figure}[ht]
    \centering
	\includegraphics[width=0.5\textwidth]{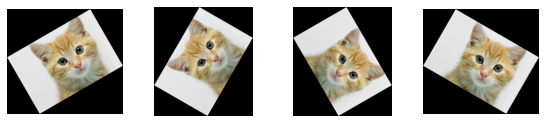}
	\caption[Augmentation Example on Cat Image]{Rotated on 4 different angles}
	\label{fig:catFig3}
\end{figure}

Geometric Augmentations effects on the shape of images, for example, rotating, zooming, resizing fig \ref{fig:catFig3}. An experiment of building a CNN model using single image has been performed in \cite{singleImage2014}. The purpose of this CNN model was to perform depth map prediction. Data augmentation techniques used in this research are random transformations of geometric augmentations, i.e., scaling the image, rotating the image in a range of -5, 5 degrees, randomly cropping the images, and flipping the images horizontally. According to the authors, if we apply 3D transformations to such images, we can achieve better accuracy. Hand gesture recognition \cite{augHandgestures}  has been proposed on a low resource data set with the use of augmentation techniques like (mirroring, reverse ordering, rotation, scaling, translation). These techniques played a vital role in increasing the dataset. \newline
\subsubsection{PhotoMetric Augmentations}
\begin{figure}[ht]
    \centering
	\includegraphics[width=0.5\textwidth]{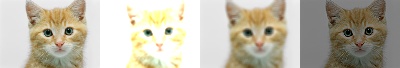}
	\caption[Augmentation Example on Cat Image]{Same image in different PhotoMetric Augmentations}
	\label{fig:catFig1}
\end{figure}

Photo can contain different colors, and these color vary because of different lightning conditions, and camera quality, a ML model can achieve increase in performance if it is trained on different lightning conditions. Photometric augmentations are the augmentations which deal with the change in color of the image e.g fig \ref{fig:catFig1} shows the image of cat at different brightness levels. H.B.Kekre, SD Thepade, V Lodha, \cite{trunctationCoding} have performed a detailed analysis using CBIR(Content-based image retrieval) techniques for image augmentations. They applied "different color spaces, Block Truncation Coding on RGB(red, green, blue) images, and Flipping the images" on 11 different categories. The comparison of augmentation techniques shown that the flipping used with all the color spaces helps improve the performance; it is stated that the YUV(composite analog signals) has the best performance among all other color spaces. 

There is always a need for extensive data for making a classifier that can outclass all other classifiers in \cite{deepImage}, but the question is where to get that much data. An increase in the dataset can be performed using several data augmentation techniques, and the performance of the model can be improved using multi-scale training and using different classifiers. Data augmentations techniques played a vital role in making the model robust for classifying different kinds of data. B. Sapp, A Saxena, and Andrew N.g proposed an idea of increasing the data set of images using green background and applying different filters on it \cite{fastData}. The proposed solution added 30\% to the accuracy of object classification; on the other hand, the dataset was increased from $10^2$ to $10^5$. This augmentation technique has decreased the effort/time for collecting datasets.\newline
\subsubsection{Simulator based Augmentations}
Images can be generated by 3D models which can generate images at several angles, this kind of data augmentation technique is known as Simulator based augmentation. The idea of image augmentation for gait analysis algorithms can be performed by the use of a simulation-based methodology for generating object-specific data \cite{gaitRecognitionAlgo}. This simulation model of an object can produce the object with different angles, different views, i.e., 3D representation gives more information and helps create more diverse views of objects. This technique has increased the dataset from 6.5 million frames to 4.5 billion frames.
\subsubsection{Combination of different Datasets}
Different datasets might have several same classes, those datasets with same classes can be combined into making another dataset for the common classes with more examples, this process of combining dataset is another form of augmentation. Two similar but different datasets PascalVoc and Microsoft COCO are combined \cite{aug2015DeepCNN} to increase the datasets, for training GoogleNet and VGG-16. The tests were performed to check the effect of results from GoogleNet, and VGG with the data separately, i.e., once on two versions of PascalVoc 2007 and 2012. The second experiment was performed after combining the two datasets, which resulted in 4 models, then two more models were trained on PASCAL(2012+2007) and Microsoft COCO separately and combined. The research concludes that an increase in dataset increases the performance; however, there are no free lunches; it comes at the cost of higher computational power. 
\newline Another scenario of combination can be using the data which is already available, for example, using data from the internet. A comparison of two different types of augmentations, i.e., geometric methods and photometric methods, are performed on the data from the internet \cite{webDataAugmentation}. Geometric methods were cropping, rotating, and flipping, whereas photometric methods were color jittering (i.e., changing color values), edge enhancement, and fancy PCA.  Those augmented datasets were then fed to a simple CNN model, and the results lead to a conclusion that cropping has shown better accuracy among all the six augmentation methods discussed here, while fancy PCA, and color jittering showed good accuracy among the photometric methods.
\subsection{Other augmentations techniques}
The in-processing part is done inside the classifier by implementing some operations on data to make it augmented. It will lead to an increase in classification performance. Some of the techniques discussed are as follows:
\subsubsection{Generative Models}
Generative models can help increasing dataset by generating more data. GAN(Generative Adversarial Networks) \cite{gans,9378518} is the example of such a model for generating images. GAN's consist of two neural networks, one for generating the new images and the second is for checking the credibility of the newly generated images.

In \cite{DeepLearning2017} a method was proposed to generate images by computing pairs of images and compared this procedure with traditional image augmentation algorithms and Generative Adversarial Networks. The results were not significant enough to approve that the new model outclassed other procedures, as the traditional model performed better than other models. But the authors proposed an idea that combining all three techniques (GAN's, affine augmentations, proposed algorithm), this combination can help to get better results. The model can be improved by applying data augmentation techniques without respect to the domains.  

Augmentation can be performed on different domains,  \cite{ratner2017learning, computers10080094}, an experiment is performed of performing augmentation on text input and image input. The procedures include making a model for generating images based on their domains and applying various affine transformations. Then the use of the Generative adversarial objective for generating images using the transformations defined. A Bayesian formulation approach is performed for applying data augmentations \cite{bayesianDataAgumentation2017}, the algorithm consists of two models, a classifier and a generator, GAN's are used as generators, and CNN's are used as classifiers. According to the experiments, the Bayesian formulations were the main contribution of this paper; those formulations improved the performance of the deep learning classification model.

GAN's can produce augmented images \cite{Diffeomorphisms}, the model first learns the data augmentation schemes and then produces the newly augmented images, which is a better approach than hand-crafting images. This model outperformed the previous way of augmenting the data using pre-defined augmentations, and the model itself learns to apply a transformation on the data.

\subsubsection{Transfer learning}
Transfer learning can help to use the knowledge of a trained model to make a new model \cite{transferlearning}. When we train a ML model, it gives us the weights, which would be later used for prediction, and those weights can be used as initial weights for the new class with fewer resources. ImageNet \cite{ImageNet} is an example of a huge dataset containing 1000 different categories, 1.2 million images, same as CIFAR and MINIST are some examples of gaining initial weights for a specific domain. The famous pre-trained models are VGG\cite{vgg}, RESNET\cite{resnet}, Inception\cite{inception}.
\subsection{Applying Augmentations}
We have defined the augmentations, but which augmentation to apply on our model is still a questions, and there are plenty of answers to this questions. Some of the methods which help us to find useful augmentations are discussed below:
\newline\newline AutoAugment is a model which is self capable of deciding which augmentation is better for the underlying problem \cite{AutoAugment} . The augmentation technique is referred to as a policy, the methodology consists of applying different augmentations on a specific dataset with some magnitudes, then classifying them to find out which augmentation technique performed better. Each policy consists of two parts: the type of augmentation method and magnitude and frequency; the results have shown that the model has performed better than the previous state-of-the-art methods. 
\newline\newline Smart augmentation is another method for finding what augmentation is useful for our dataset, and model \cite{smartAugmentation}. Smart Augment trains a model on domain specific augmentations at which the network loss is reducing. The author used a faces dataset from 5 different sources, then performed the smart augmentation algorithm of generating images using a network of networks, and then performed a gender classification task. Some other examples are KeepAugment \cite{Gong_2021_CVPR}, and RandAugment \cite{Cubuk_2020_CVPR_Workshops}.

\section{Methodology}\label{sec:methods}
\begin{figure}[ht]
    \centering
	\includegraphics[width=0.35\textwidth]{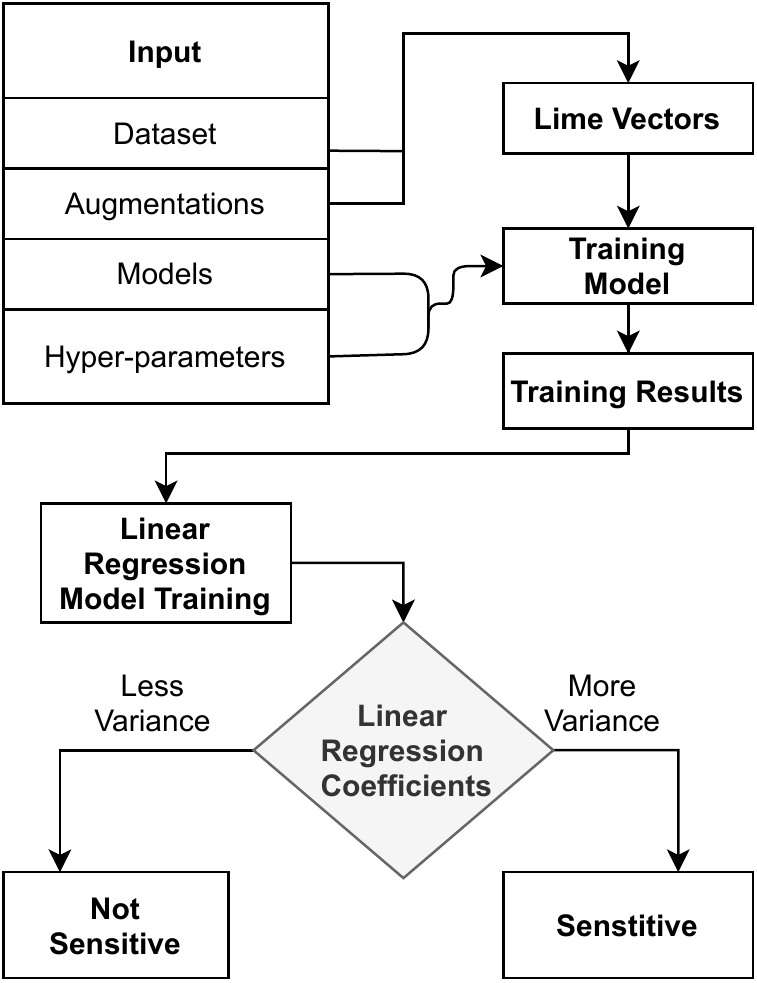}
	\caption[Lime analysis]{Methodology}
	\label{fig:methodology}
\end{figure}

To study the effect of hyper-parameters on augmentations, we are using lime analysis and linear regression coefficients, so to find the effect we defined hyper-parameters and augmentations for our experiments which are discussed below. To test our hypothesis we came up with a methodology, which can be visualized in fig-\ref{fig:methodology}, we used the combination of augmentation based on lime vectors which later applied to dataset for training. The combination of hyper-parameter and the machine learning models were also defined. We trained each model with specific hyper-parameter on the dataset with specific augmentation, and collected all the results. 
\newline
The results of training can be visualized in table-\ref{tab:vectable}, we have a combination of hyper-parameter, and lime vector, which maps to accuracy and loss. These results were later fed to the linear regression model, this linear regression model resulted into coefficients of each augmentations. Those coefficients were used to determine the relation between augmentations and hyper-parameters. 
The main components of our methodology are as follows:
\subsection{Machine Learning Model}
To deal with image classification task we used two versions of an image classification model ResNet(Residual Networks) \cite{He2016resNet}, which are ResNet50 and ResNet101.

ResNet50 is a strong model with 48 convolution layers, 1 max pool layer, 1 average pooling layer, which results into 25.6 million trainable parameters \cite{zagoruyko2017wide}. 

ResNet101 is just an upgraded version of ResNet50 with more layers, in ResNet101 there are 101 layers, and it has 44.5 million trainable parameters \cite{zagoruyko2017wide}. 

Our second model was linear regression, which is used for finding the regression coefficients of each augmentations. These coefficients later define the importance of each augmentation.
\subsection{HyperParameters}
It's a known fact that hyper-parameters play an essential role in the performance of the model. In our experiment we are checking the effect of hyper-parameters, for that purpose we defined the four different combination of optimizers, and epochs, i.e SGD with (20/15 epochs), Adam with (20/15 epochs).

\subsection{Data}
\subsubsection{Dataset}
To perform the experiment, we have an online available dataset FashionMNIST. This dataset consists of images from 10 different fashion products i.e. (T-shirt, Trouser, Pullover, Dress, Coat, Sandals, Shirt, Sneakers, Bag, and Ankle boots), all of the images are in gray-scale and of size 28x28. Each of the category is treated as a class, and each class has 7000 images in the dataset, leading to 70,000 images in total \cite{xiao2017fashion}. Initially, there are 60,000 images in training set and 10,000 images in testing set, we have modified the training set and created a validation set from it, our training set consist of 51,000 images and validation set consists of 9000 images.
\subsubsection{Augmentations}
Albumentation is a library with 70 different augmentations, which are fast and easy to apply  \cite{info11020125}; Our dataset has grey scale images, so we have utilized the augmentations which are suitable for our dataset, also the augmentations were applied with default parameters. Our chosen augmentations are \emph{Transpose}, \emph{Blur}, \emph{Downscale}, \emph{Equalize}, \emph{GaussNoise}, \emph{GaussianBlur}, \emph{InvertImg}, \emph{ShiftScaleRotate}, \emph{RandomRotate90}. 

\subsection{Lime Analysis}
Lime(Local Interpretable Model-Agnostic Explainations) is a tool which helps to understand the behavior of the model, the assumptions made by lime are based on the prediction of a model \cite{ribeiro2016lime}. Lime can be used for interpreting tabular data, text data, and images, let's see an example of text interpretation. A model is trained to find the spam emails, so email contains text, now a trained model will analyze the text to give us what text is spam. A text is combination of words, but we don't know what word or combination of words leads to spam. Lime helps to find what words, combination of words are effecting the decision. Lime follows a simple mechanism, it will take a sentence which is considered as spam by a model, and make some samples of the sentence. Each sample will contain some part of sentence with some words removed, the recorded predictions for each sample will tell us how much each word is contributing to the model for prediction of spam. 

We have used such an implementation of Lime for our methodology, as we have several augmentations, and to know which augmentation is weighing how much we used Lime. We have nine different augmentations, so we created different vectors of length nine, each index is representing an augmentation and the value is either 0's or 1's. 0 indicates that there is absence of such augmentation, and 1 means we have applied this augmentation. 

We have created 28 such vectors which were than applied to our dataset, and then we trained our model, as we are dealing with different hyper-parameters, so we trained our model for each vector for four different times, then stored the accuracy and loss for each model on test data. An example of two such vectors can be seen in table \ref{tab:vectable}, in vector one \emph{transpose, GuassianNoise, GuassianBlur, ShiftScaleRotate} and \emph{RandomRotate90} is applied to dataset for training, which resulted in terms of accuracy and loss for the model, and respectively for Vector 2. 

After training all the vectors we got accuracy, loss of each vector as presented in table \ref{tab:vectable}, that resulted into 224 samples, i.e 28 samples for each model with a specific hyper-parameter. We trained two(accuracy, loss) linear regression models for every combination of model and hyper-parameter, and recorded their regression coefficients. These regression coefficients are the indicator of effect of each augmentation on the model. We will discuss the regression coefficients more in next section of results \ref{sec:results}. 

\begin{table}[ht]
\centering
\caption{Dataset with applied augmentations and their respective performance. In table $a_{i}$ represents the augmentation, where $i$ is the number of augmentation which is applied}
\begin{tabular}{ll|l|l|l}
                                    & \multicolumn{2}{|l|}{ResNet50} & \multicolumn{2}{l}{ResNet101} \\ \hline
\multicolumn{1}{l|}{Augmentations}  & Acc           & Loss          & Acc            & Loss          \\ \hline
\multicolumn{1}{l|}{$a_{2},a_{5},a_{7},a_{9}$}    & 87.2          & 34.4          & 84.9           & 43.7          \\ \hline
\multicolumn{1}{l|}{$a_{1},a_{5},a_{6},a_{8},a_{9}$} & 87.8          & 33.3          & 84.8           & 40.2         
\end{tabular}
\label{tab:vectable}
\end{table}

\section{Results} \label{sec:results}
The following combinations of hyper-parameters were used with two different models, for testing their effect on augmentations: SGD with (20/15 epochs), Adam with (20/15 epochs).

The example of lime vectors can be visualized in table \ref{tab:vectable}, for this case we tested the outcome of 28 such vectors.

After defining the model hyper-parameters, the models were trained on different vectors, and their accuracy and loss are recorded for each hyper-parameter; an example of accuracy and loss against vectors can be visualize in table \ref{tab:vectable}. A dataset of 28 different vectors is created with four results, i.e. (accuracy, loss for each model), and on each hyper-parameter; it led to a dataset of $2(models)*4(hyper-parameters)*28(vectors)$ instances, which means a total of 224 different models were trained with different augmentation combinations. These models resulted into accuracy and loss, that were later used to train the linear regression model.

After training the linear regression models on the results of all the vectors, the calculated Intercept for each model (i.e., ResNet50, ResNet101) and the corresponding comparison matrix (i.e., Accuracy, Loss) is as follows:

\begin{tikzpicture}[scale=0.75]
\begin{axis}[
    title={Linear Regression model Intercept distribution},
    xlabel={Hyper Parameters},
    ylabel={Intercept},
    xmin=0, xmax=5,
    ymin=-250, ymax=1400,
    xtick={1,2,3, 4},
    ytick={-165, 0,90, 240,480,960,1200},
    legend pos=north west,
    ymajorgrids=true,
    grid style=dashed,
]

\addplot[
    color=blue,
    mark=square,
    ]
    coordinates {
    (1,90.512)(2,89.323)(3,89.3675)(4,89.3883)
    };
\addplot[
    color=red,
    mark=square,
    ]
    coordinates {
    (1,33.3264)(2,34.0535)(3,29.5997)(4,33.64966)
    };
\addplot[
    color=green,
    mark=square,
    ]
    coordinates {
    (1,90.3598)(2,91.2862)(3,91.54377)(4,71.6030)
    };
\addplot[
    color=black,
    mark=square,
    ]
    coordinates {
    (1,42.5369)(2,-163.4952)(3,116.7092)(4,1255.60897)
    };
    \legend{ResNet50 Accuracy,ResNet50 Loss, ResNet101 Accuracy, ResNet101 Loss}
    
\end{axis}
\label{fig:intercept}
\end{tikzpicture}

To understand the effect of augmentation, the linear regression coefficients of each augmentation were plotted. There were two different comparison metrics, leading to two subsets of plots: i.e. accuracy and loss.

In the plot x-axis represents the augmentation i.e. the id of each augmentation i.e. 0-\emph{Transpose}, 1-\emph{Blur}, 2-\emph{Downscale}, 3-\emph{Equalize}, 4-\emph{GaussNoise}, 5-\emph{GaussianBlur}, 6-\emph{InvertImg}, 7-\emph{ShiftScaleRotate}, 8-\emph{RandomRotate90}. 

In the plot Y-axis represents the regression coefficient for each augmentation. In the following plots blue line refers to 20 epochs with SGD, the red line refers to SGD with 15 epochs, the green line represents the models' performance trained on Adam with 20 epochs, and black line for the performance of the model with a hyperparameter of Adam as optimizer with 15 epochs. 
\subsection{Accuracy Plots}

\begin{tikzpicture}[scale=0.75]
\begin{axis}[
    title={ResNet50 Normalized},
    ylabel={Coefficient},
    ymin=-2.5, ymax=2.5,
    xtick={0,1,2,3,4,5,6,7,8},
    ytick={-2.0,-1.6,-1.2, -0.5,0,0.5,1.2,1.8},
    legend pos=north west,
    ymajorgrids=true,
    grid style=dashed,
]

\addplot[
    color=blue,
    mark=square,
    ]
    coordinates {
    (0,-1.37753) (1,-0.38749) (2,-1.09627) (3,0.82165) (4,1.00407) (5,1.6181) (6,-0.90568) (7,0.69845) (8,-0.3753)};
\addplot[
    color=red,
    mark=square,
    ]
    coordinates {
    (0,-1.05912) (1,-0.70609) (2,0.24839) (3,0.06388) (4,0.37245) (5,0.85112) (6,-1.98115) (7,0.90196) (8,1.30855)};
\addplot[
    color=green,
    mark=square,
    ]
    coordinates {
    (0,-1.57819) (1,-0.98549) (2,-0.6173) (3,0.83706) (4,1.54998) (5,1.21236) (6,-0.64639) (7,0.37885) (8,-0.15087)};
\addplot[
    color=black,
    mark=square,
    ]
    coordinates {
    (0,-1.09506) (1,-1.8638) (2,0.85154) (3,1.03695) (4,0.59073) (5,0.12869) (6,-0.78837) (7,-0.09767) (8,1.23698)};
    
\end{axis}
\label{fig:restnet50norm2Acc}
\end{tikzpicture}

\begin{tikzpicture}[scale=0.75]
\begin{axis}[
    title={ResNet101, Normalized},
    ylabel={Coefficient},
    ymin=-2.5, ymax=2.5,
    xtick={0,1,2,3,4,5,6,7,8},
    ytick={-2.0,-1.6,-1.2, -0.5,0,0.5,1.2,1.8},
    legend pos=north west,
    ymajorgrids=true,
    grid style=dashed,
]

\addplot[
    color=blue,
    mark=square,
    ]
    coordinates {
    (0,-0.66805) (1,-1.30232) (2,0.21443) (3,1.30395) (4,1.64193) (5,0.89983) (6,-0.38051) (7,-0.86491) (8,-0.84435)};
\addplot[
    color=red,
    mark=square,
    ]
    coordinates {
    (0,-0.09396) (1,-2.21461) (2,0.74823) (3,0.64976) (4,-0.62531) (5,-0.16057) (6,0.92741) (7,-0.49069) (8,1.25975)};
\addplot[
    color=green,
    mark=square,
    ]
    coordinates {
    (0,-1.00143) (1,-1.28967) (2,1.46033) (3,0.42909) (4,0.97253) (5,0.83223) (6,-1.49945) (7,0.29897) (8,-0.20261)};
\addplot[
    color=black,
    mark=square,
    ]
    coordinates {
    (0,-0.11416) (1,-0.83631) (2,0.41451) (3,1.22928) (4,-0.96455) (5,2.08849) (6,-0.63685) (7,-0.9144) (8,-0.266)};
    
\end{axis}
\label{fig:restnet101norm2Acc}
\end{tikzpicture}
\subsection{Loss Plots}
\begin{tikzpicture}[scale=0.75]
\begin{axis}[
    title={ResNet50 Normalized},
    ylabel={Coefficient},
    ymin=-2.5, ymax=2.5,
    xtick={0,1,2,3,4,5,6,7,8},
    ytick={-2.0,-1.6,-1.2, -0.5,0,0.5,1.2,1.8},
    legend pos=north west,
    ymajorgrids=true,
    grid style=dashed,
]

\addplot[
    color=blue,
    mark=square,
    ]
    coordinates {
    (0,1.86404) (1,0.67803) (2,0.58226) (3,-0.60635) (4,-0.99853) (5,-1.567) (6,0.77833) (7,-0.49495) (8,-0.23582)};
\addplot[
    color=red,
    mark=square,
    ]
    coordinates {
    (0,1.46457) (1,0.09303) (2,-0.96806) (3,0.6726) (4,-0.55259) (5,-0.52453) (6,1.6835) (7,-1.31929) (8,-0.54922)};
\addplot[
    color=green,
    mark=square,
    ]
    coordinates {
    (0,1.85244) (1,1.27735) (2,0.00981) (3,-0.62812) (4,-1.5652) (5,-1.00633) (6,0.24856) (7,-0.10154) (8,-0.08696)};
\addplot[
    color=black,
    mark=square,
    ]
    coordinates {
    (0,0.41434) (1,1.7776) (2,-1.30983) (3,-0.41325) (4,-0.10486) (5,-0.19157) (6,0.95644) (7,0.48017) (8,-1.60903)};
    
\end{axis}
\label{fig:restnet50norm_loss_2}
\end{tikzpicture}

\begin{tikzpicture}[scale=0.75]
\begin{axis}[
    title={ResNet101 Normalized Loss},
    ylabel={Coefficient},
    ymin=-2.5, ymax=2.5,
    xtick={0,1,2,3,4,5,6,7,8},
    ytick={-2.0,-1.6,-1.2, -0.5,0,0.5,1.2,1.8},
    legend pos=north west,
    ymajorgrids=true,
    grid style=dashed,
]

\addplot[
    color=blue,
    mark=square,
    ]
    coordinates {
    (0,-0.91805) (1,0.4217) (2,0.33034) (3,-2.01824) (4,-0.58406) (5,-0.35578) (6,0.94881) (7,1.26523) (8,0.91005)};
\addplot[
    color=red,
    mark=square,
    ]
    coordinates {
    (0,0.63276) (1,0.98258) (2,-1.247) (3,-0.58474) (4,-0.23721) (5,1.98876) (6,-0.99077) (7,0.2737) (8,-0.81808)};
\addplot[
    color=green,
    mark=square,
    ]
    coordinates {
    (0,-0.5953) (1,0.83343) (2,-1.7796) (3,1.25519) (4,0.04326) (5,0.66887) (6,0.90186) (7,-1.39342) (8,0.06571)};
\addplot[
    color=black,
    mark=square,
    ]
    coordinates {
    (0,-1.27931) (1,-0.76738) (2,0.88565) (3,0.21956) (4,-0.4324) (5,1.62897) (6,-0.5824) (7,1.32741) (8,-1.0001)};
    
\end{axis}
\label{fig:restnet101normloss2}
\end{tikzpicture}

Two criteria for measuring the sensitivity of an augmentation were introduced, which are consistency, and influence.

\textbf{Consistency}:
To find the consistency, first step is to calculate the variance of the coefficients of augmentations across all the hyper-parameters and for each model. In this case there were two models, so first the variance of coefficients for each model is calculated, and than the average variance of both models. This accumulated variance is directly proportional to the consistency of an augmentation i.e. if average variance increases the consistency decreases and vice versa. 

The following equation explains the consistency, such that $ \sigma_{v}^{2} $ is the variance of $_{k}$ models against all hyper-parameters $l$,  and for each augmentation $i$.
\begin{equation}
\centering
    \label{equationvar}
    \begin{multlined}
    sensitivity_{i} = \frac{1}{k}\sum_{v=0}^{k}\sigma_{v} ^{2}\\
    consistency_{i}= \frac{1}{sensitivity_{i}}
    \end{multlined}
\end{equation}

After variance is calculated for each augmentations, the less the variance the less the sensitivity. The threshold criteria was, the augmentation with sensitivity value less than 0.2 are non-sensitive i.e in this case the variance for augmentations 1,4,8 were less than 0.2, and also from the graphs we can see that these augmentations are barely varying, in both models, so these augmentations are consistent/non-sensitive. An example of sensitivity can be visualised from table \ref{tab:sensitivity}.

\begin{table}[ht]
\centering
\caption{The sensitivity of augmentations is dependent upon the variance of coefficients of augmentation across all of it's hyper-parameters.}
\label{tab:sensitivity}
\begin{tabular}{|c|c|c|c|}
\hline
Aug & VarM\_\{1\} & VarM\_\{2\} & Sensitivity \\ \hline
1   & 0.060465    & 0.196556    & 0.128511    \\ \hline
2   & 0.402366    & 0.334197    & 0.368281    \\ \hline
3   & 0.758516    & 0.299119    & 0.528818    \\ \hline
4   & 0.183786    & 0.185314    & 0.184549    \\ \hline
5   & 0.268516    & 1.566889    & 0.917703    \\ \hline
\end{tabular}
\end{table}

\textbf{Influence}:
We know that linear regression coefficients have signs, i.e. negative signs means that augmentation is going to effect the model negatively, and vice versa. To find reliability, there is a need to check what is the effect of an augmentation, this effect is defined as influence.

The influence is calculated by averaging the coefficients of an augmentation across all the hyper-parameters, for each model, and than averaging the average as shown in fig below:
\begin{equation}
    \label{equationvar}
    \begin{multlined}
    avg_{k}=avg(coef_{k})\\
    avg_{i}=\frac{1}{k}\sum_{t=0}^{k} (avg_{t})\\
    influence_{i} = avg_{i}
    \end{multlined}
\end{equation}

\textbf{Reliability}:
We argue that if an augmentation is reliable it has to be both consistent and less influenced, so the consistency and influence of an augmentation is multiplied to result into combined effect that defines the reliability of augmentation. The reliability recognizes the effect of a specific augmentation on the performance of a model.
\begin{equation}
    \label{equationvar}
    \begin{multlined}
    reliability_{aug_{i}}=consistency_{i}*influence_{i}
    \end{multlined}
\end{equation}
After calculating the reliability of each augmentation, we state the threshold for identifying the augmentation as reliable or non-reliable, in this case it is sated that the top 3 values with larger positive coefficients are reliable.

\begin{table}[ht]
\centering
\caption{This table shows that the reliability of an augmentation, how the augmentation that is applied to the ML model is going to affect the performance of the model.}
\label{tab:reliabilityTable}
\begin{tabular}{|l|l|r|l|}
\hline
Aug & Sensitivity & \multicolumn{1}{l|}{Influence} & Reliability \\ \hline
1   & 0.128511    & -0.873437                      & -0.112246   \\ \hline
2   & 0.368281    & -1.198223                      & -0.441283   \\ \hline
3   & 0.528818    & 0.277983                       & 0.147002    \\ \hline
4   & 0.184549    & 0.796453                       & 0.146985    \\ \hline
5   & 0.917703    & 0.567729                       & 0.521006    \\ \hline
\end{tabular}
\end{table}

\section{Discussion}
To identify the sensitivity a criteria is already set up, that is based on consistency, in this case we have stated a threshold of 0.2 for sensitivity i.e the augmentations with sensitivity more than threshold are sensitive and vice versa. Based on this criteria it is discovered that augmentations Transpose, Equalize, and ShiftScaleRotate were insensitive to the hyperparameters, while augmentations GaussianBlur, InvertImg and RandomRotate90 are most sensitive to the hyperparameters.

\begin{figure}[ht]
    \centering
	\includegraphics[width=0.20\textwidth]{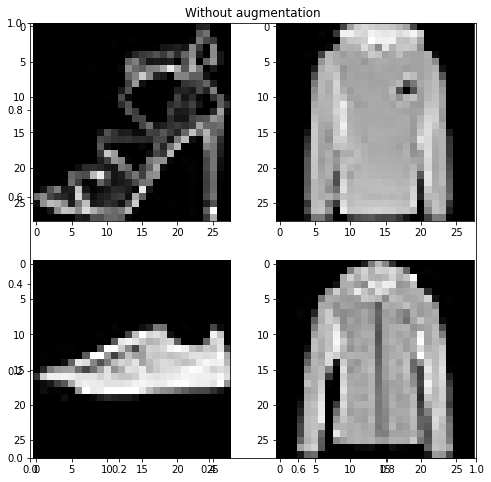}
	\includegraphics[width=0.20\textwidth]{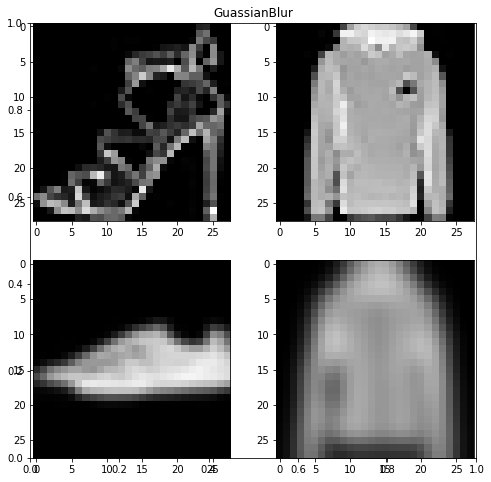}
	\caption[GaussBlur]{No augmentation vs Gaussian Blur}
	\label{fig:gaussBlur}
\end{figure}

\textit{GaussianBlur} when applied it returns the blurred image with random kernel size applied on. It can be visualized from the graphs that \textit{GaussianBlur} is giving us positive coefficients for all the hyperparameters when the model is ResNet50, and $\frac{3}{4}$ positive coefficients for ResNet101. But the consistency for this augmentation is varying with lot's of difference, i.e. augmentation is sensitive to hyper-parameter, the example of this augmentation applied to dataset can be visualized from fig \ref{fig:gaussBlur}

\begin{figure}[ht]
    \centering
	\includegraphics[width=0.20\textwidth]{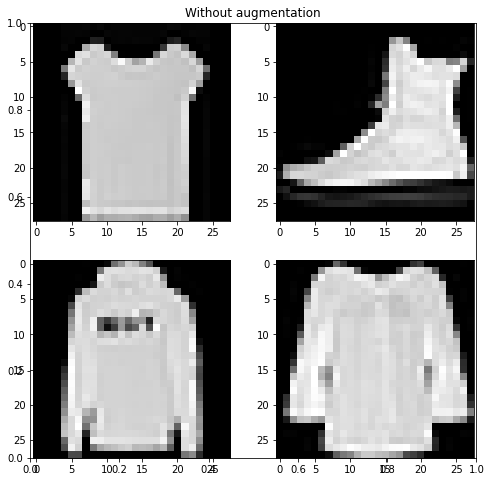}
	\includegraphics[width=0.20\textwidth]{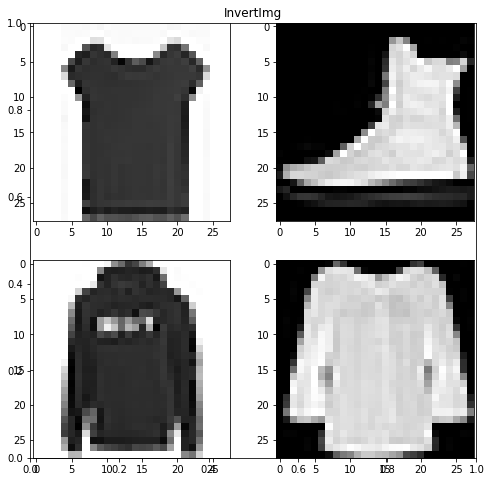}
	\caption[invertImg]{No augmentation vs Invert Image}
	\label{fig:invertImg}
\end{figure}

The augmentation \textit{InvertImage} takes a parameter of probability for applying the augmentation, the default value for this augmentation is $0.5$ which means that the transform will apply augmentation to half of the images, fig \ref{fig:invertImg} shows the resultant images when this augmentation was applied.

\begin{figure}[ht]
    \centering
	\includegraphics[width=0.20\textwidth]{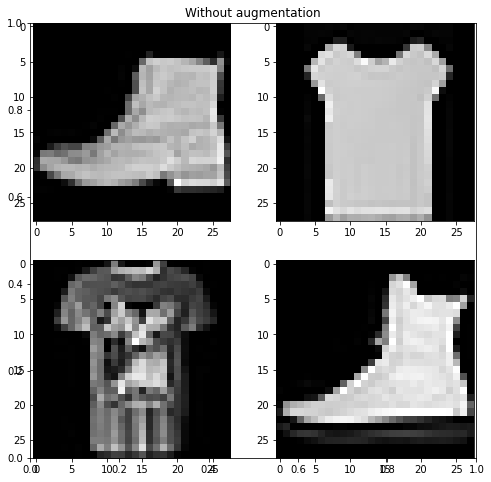}
	\includegraphics[width=0.20\textwidth]{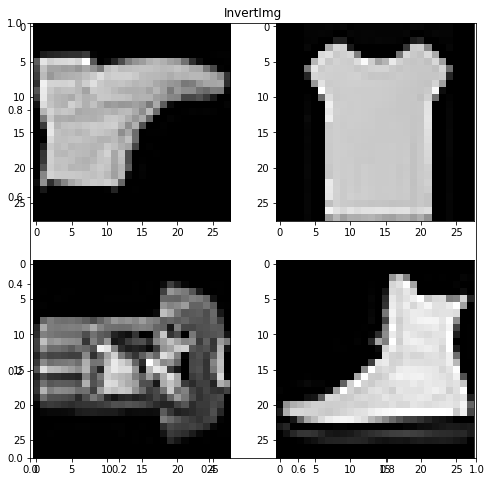}
	\caption[rotateImg]{No augmentation vs RandomRotate90}
	\label{fig:rotateImg}
\end{figure}

\textit{RandomRotate} augmentation applies the geometric transformations of rotating the image with multiples of $90^{\circ}$  randomly with a probability, we set this probability to default i.e. 0.5, fig \ref{fig:rotateImg} shows the results when this augmentation was applied. 

The above figures were not giving enough explanation about the reason of an augmentation being sensitive, so we explored the augmentation which were not sensitive, and tried to find out what factors are providing resistance to the sensitivity.

\begin{figure}[ht]
    \centering
	\includegraphics[width=0.20\textwidth]{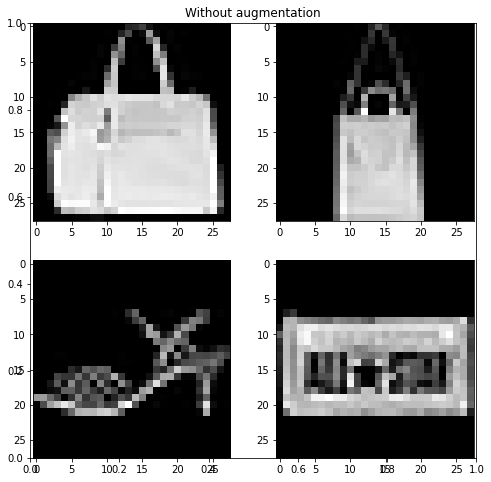}
	\includegraphics[width=0.20\textwidth]{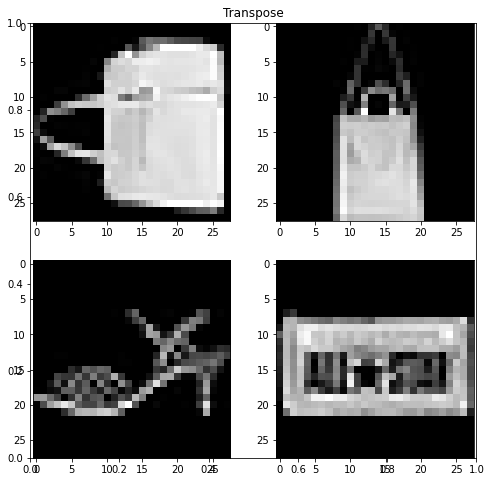}
	\caption[rotateImg]{No augmentation vs Transpose}
	\label{fig:transpose}
\end{figure}

The augmentation \textit{Transpose} is applied with a probability of $0.5$ and the result of this augmentation can be visualized from fig \ref{fig:transpose}. This augmentation transposed the images and changed the shape of an image, the difference is not much significant, also a point to consider here is, that this augmentation is geometric augmentation.

\begin{figure}[ht]
    \centering
	\includegraphics[width=0.20\textwidth]{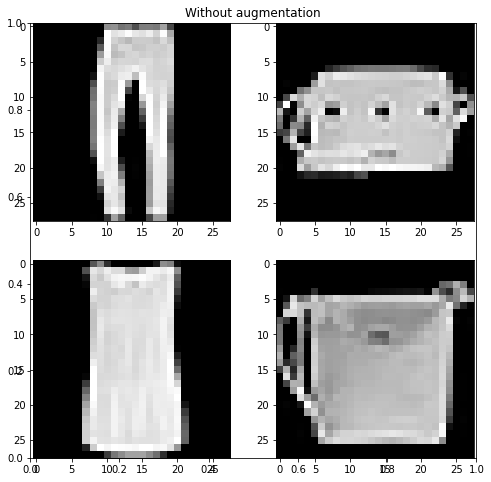}
	\includegraphics[width=0.20\textwidth]{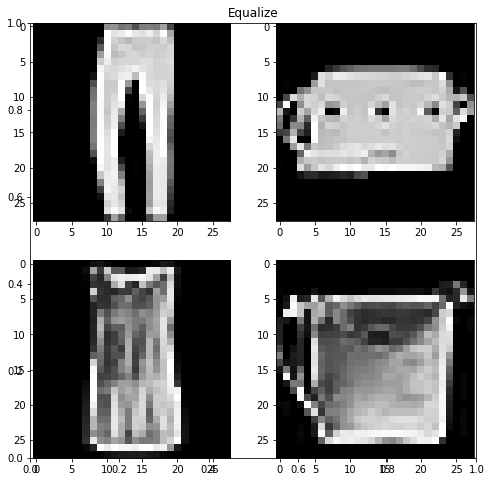}
	\caption[equalize]{No augmentation vs Equalize}
	\label{fig:equalize}
\end{figure}

\textit{Equalize} augmentation is applied with the probability of $0.5$ and the images generated using this augmentation are presented in fig.\ref{fig:equalize}, this augmentation equalizes the histogram of the image to gives more smoothed details to the image, this augmentation is photo-metric augmentation for equalizing the distribution.

\textit{ShiftScaleRotate} applies a transform of three different geometric augmentations (i.e. shifting, scaling and rotating) we have used the default values and a probability of $0.8$. This is also a geometric augmentation, but with a slight change in shapes.
\begin{figure}[ht]
    \centering
	\includegraphics[width=0.20\textwidth]{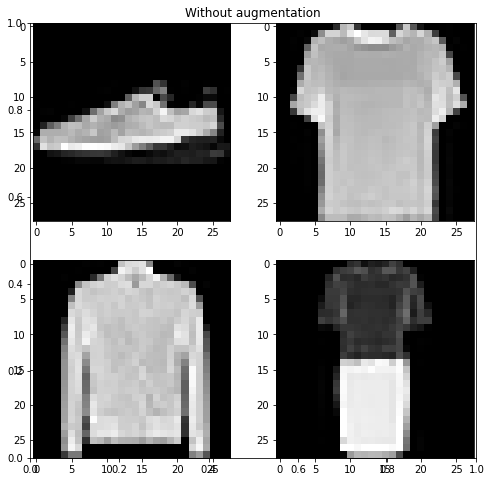}
	\includegraphics[width=0.20\textwidth]{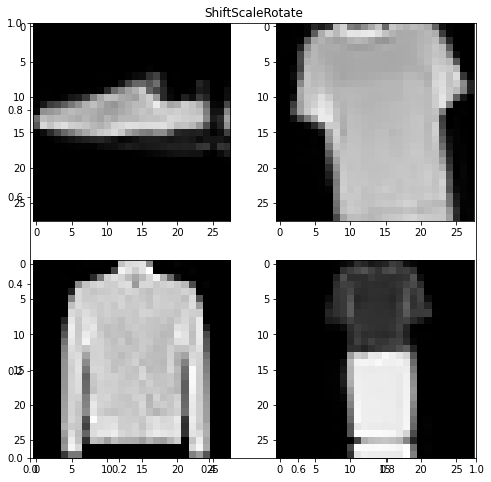}
	\caption[shiftscalerotate]{No augmentation vs Shift Scale Rotate}
	\label{fig:shiftScalerotateImg}
\end{figure}

After observing the sensitive and non-sensitive augmentations, a trend can be seen that, the augmentations applied with smaller intensities are less effected by the hyper-parameters. To elaborate it more, let's discuss the non-sensitive augmentations, we have applied \textit{transpose}, it is not making a significant change in an image, except changing the rows and columns, \textit{equalize} is also re-distribution the intensities which is not a significant intensity, and \textit{shift-scale-rotate} is changing the shape of image with a lower limits, i.e. \textit{shift\_limit=$0.0625$, scale\_limit=$0.1$, rotate\_limit=$15^{\circ}$}.

In contrast \textit{Gaussian Blur} is completely changing the shape of image, and it has blurred images that are not even classifiable by naked eye, \textit{Random rotate} is rotating the image at multiples of $90^{\circ}$ degrees, which is again changing the shape of image by huge intensity, while Invert Image is completely changing the color distribution of an image which is again an indicator of high intensity.

We accept the null hypothesis that some augmentations are sensitive to the hyper-parameters, and the reason of this sensitivity is the intensity of the applied augmentation, if the augmentation is applied with huge intensity, it will led to changing effect for each hyper-parameter.

\section{Conclusion}
In this research, a simple technique for finding the effect of hyper-parameters on Augmentations is proposed, this technique identified which augmentations are sensitive to the hyper-parameters for FashionMNIST Dataset. Our research has shown that there are augmentations which are sensitive to the hyper-parameters because of the intensity of augmentation. Also we proposed a methodology to find the reliability of an augmentation, which is going to help the decision maker for choosing the augmentation that best suits the model and dataset given some hyper-parameters. The reliability shows whether the augmentation is going to effect the performance of model positively or negatively. 

Our future work will consist of working on this methodology to determine the number of vectors, that are required for the model to give the best augmentations and tuning the data for observing our model's performance on different sizes of the dataset. 

\printbibliography

\end{document}